# STREAMLINED SHAPE OF CYBORG COCKROACH PROMOTES TRAVERSABILITY IN CONFINED ENVIRONMENTS BY GAP NEGOTIATION

Kazuki Kai, Le Duc Long, and Hirotaka Sato*

School of Mechanical & Aerospace Engineering, Nanyang Technological University; 50 Nanyang Avenue, 639798, Singapore

*Corresponding author. Email: hirosato@ntu.edu.sg

**Abstract**

The centimeter-scale cyborg insects have a potential advantage for application in narrow environments where humans cannot operate. To realize such tasks, researchers have developed a small printed-circuit-board (PCB) which an insect can carry and control it. The electronic components usually remain bare on the board and the whole board is mounted on platform animals, resulting in uneven morphology of whole cyborg with sharp edges. It is well known that streamlined body shape in artificial vehicles or robots contributes to effective locomotion by reducing drag force in media. However, little is known how the entire body shape impacts on locomotor performance of cyborg insect. Here, we developed a 10 mm by 10 mm board which provided electrical stimulation via Sub-GHz communication and investigated the impact of physical arrangement of the board using Madagascar hissing cockroach. We compared the success rate of gap negotiation between the cyborg with mounted board and implanted board and found the latter outperformed the former. We demonstrated our cyborg cockroach with implanted board could follow faithfully to the locomotion command via antennal or cercal stimulation and traverse a narrow gap like air vent cover. In contrast to the conventional arrangement, our cyborg insects are suitable for application in a concealed environment.

**Introduction**

Insects are often utilized as a platform for cyborg animals due to their small size, ability to avoid obstacles and locomotor capability. They can attach to a rough texture like wood trunk with claws, while the pad between the claws, called arolium, is used to attach to a slippery surface such as glass or acrylic plate. Also, insects receive input from surrounding through sensory organs and respond to it. As cyborg insects can exploit such ability of the platform animal, they have potential advantages for application in complex and unpredictable environments.

Locomotion control is a fundamental element of cyborg insects as they are required to navigate to the desired place in real application. To accomplish this, researchers have developed stimulation protocols using electrical current delivered to various parts of insect body. Electrical current causes action potentials in muscle or neurons nearby, therefore it can drive the motor or nervous system to induce locomotion. In cockroaches and beetles, for example, electrical stimulation to left or right antenna induces turning to the other side. A flying beetle changes its direction or attitude when the current is directly passed to the flight muscle. Electrical stimulation to the ventral nerve cord, which interconnects the brain and the motor system in the thorax, can also bias the walking direction of cockroach. Recent studies show that polarity and temporal pattern of electrical pulse train affect the efficiency of stimulation. Liu et al. [1] revealed that biphasic current pulse outperforms monophasic pulse in terms of stimulus effectiveness and sustainability. On the other hand, Nguyen et. al. demonstrated that burst stimulation can induce stable reactions compared to continuous stimulation. Electrical stimulation can achieve not only manipulation of locomotor parameters (forward and angular velocity, flight attitude) but also induction of different modes of locomotion. By changing the stimulation pattern among the leg muscles in beetle, Cao et al. selectively induced galloping or tripod gait [3]. When a locust jumps, the muscles of hindleg are activated with a specific timing and electrical stimulation can reproduce jumping. In real application, such locomotion control must be carried out with a wireless printed circuit board which is small enough for insects to carry.

In recent years, electronic components have been miniaturized so that a small PCB can house various sensors to realize useful functions besides the microcontroller and digital-to-analog converter for stimulation. Localization is one of the most important functions in cyborg insects for urban search and rescue missions (USAR) to locate a victim. However, Global Positioning Systems (GPS) are not feasible as GPS signals may be unavailable under the rubble. Instead, Cole et al. achieved centimeter-level accuracy using on-board inertial measurement unit [4]. Cameras are powerful tools to discover a living human during USAR. Tran-Ngoc et al. developed a cyborg cockroach which is equipped with an infrared camera and demonstrated on-board real-time human detection [5]. An RGB camera was deployed for path planning or obstacle avoidance. A simple time-of-flight (TOF) distance sensor can also provide a solution for the trouble derived from the platform animal. Cockroaches have a natural tendency to go into a dark and narrow space such as the corner of the arena and settle down that are not desirable for cyborg operation. A TOF sensor installed in front of cyborg cockroach detects the presence of the corner to initiate stimulation to escape from such condition. Furthermore, Kakei et. al. developed a thin flexible solar cell which can cover the abdomen of Madagascar hissing cockroach to improve the power consumption of the system [6]. In any case, electronic components are exposed on the board, and the board and battery are mounted on the insect without any shield. So far, little is known about the physical constraints of the board used in cyborg cockroach.

It is well known that streamlined body shape of animals and artificial vehicles including aircraft, ship, and submarine contributes to efficient movement in media by reducing fluid drag. Similarly, body shape plays an important role in terrestrial animals such as cockroaches and beetles. These insects move through the cluttered terrains in nature where they are required to negotiate obstacles like grass, sticks, shrubs, rotten trunk of trees. Cockroaches showed high traversal performance in a narrow grass-like beam whereas reduction in body roundness by additional artificial shell led to decreased traversal performance due to the increased resistance from environments. In conventional assembly of cyborg insects, the electronics are mounted as it is, leading to the reduction of roundness of whole body. This suggests the possibility of degraded traversal performance of cyborg insects.

One of the most important applications of cyborg insects is USAR in a concealed space at post disaster sites, which is inaccessible to humans. In such situations, cyborg insects need to traverse cluttered environments where artificial obstacles such as concrete slabs, power/signal cables, and air vent may block the cyborg. In this paper, we constructed a small board for locomotion control and developed a surgical protocol to implant the board into the roach's body to minimize the change in body shape. As the body remained rounded, the cyborg cockroach could be controlled to navigate through a narrow gap like an air vent.

## 2. Methods

### 2.1. Animals

We used adult male Madagascar hissing cockroaches for all experiments. The cockroaches were reared in an individually ventilated cage (NexGen IVC 500, Allentown) and fed with carrots. The temperature and humidity of each cage were kept at 25 °C and 60%, respectively. To keep the animals in healthy condition, the cages were cleaned once a week.

### 2.2 Printed circuit board for electrical stimulation

A wireless printed circuit board (PCB) was designed and constructed to deliver electrical stimulation to cyborg cockroaches (Fig. 1). A microcontroller unit (MCU: CC1310F128) was selected as the main core for its low power consumption and long-range communication through obstacles. The MCU served wireless communication with the central board through an embedded sub-GHz module and controlled a 12-bit digital-to-analog converter chip (DAC: AD5624). The board provided 4 isolated stimulation channels through the DAC to generate biphasic electrical stimulation. The board measured 10 mm long, 10 mm wide, and 3.5 mm high, weighing 0.5 g. A 9mAh lithium-polymer battery (11.7 mm in length, 7 mm in width and 3 mm in height) was used to empower the board.

### 2.3. Obstacle track

To examine the traversal performance of the cockroaches to negotiate a narrow gap, we built a custom-made horizontal track with a movable shutter (Fig. 2A). The track measured 20 cm long by 4.7 cm wide, and both sides of the track were enclosed by a wall 5.0 cm high. A linear motion guide was attached at 12.5 cm from the entrance, and a 3D-printed shutter was mounted to the stage on the motion guide. A counterweight was connected to the stage with a nylon string and hung on the other side through a pulley installed on the top. The apparatus allowed precise control of the gap height and the force necessary to lift the shutter. The gap was set to 8 mm and the total weight of the movable shutter was 146.5 g, whereas the counterweight weighed 96.5 g. The track was open to above for 10.5 cm from the entrance (before the shutter), and the rest was shaded by a cover to darken the track so that the cockroaches were encouraged to approach the shutter. A sheet of sandpaper (60 grit) was fixed to the track floor using double-sided tape to provide sufficient friction to the insects.

### 2.4. Arrangement of board and battery

To investigate how the body shape of cyborg insect impacts on the traversal performance, we compared two arrangements of the board and battery: mounted on top and implantation (Fig. 2 and Supplementary Fig. 1).

For mounted arrangement, a thin plastic plate was attached to the mesothorax. The surface of the cuticle was roughened using a Dremel tool and the plastic plate was secured with superglue. The board and battery were arranged on the plate along the longitudinal axis of the insect body. The bottom side of the board was uneven due to the electronic components, so we used thick double-sided tape to ensure tight bonding between the plate and the board.

A piece of double-sided tape was used to attach a battery to the plate, too. The height of the board set was 4 mm including the plate.

For implantation, the board and the battery were coated with silicone elastomer (Sylgard 184, Dow) for electrical insulation. The cockroaches were anesthetized using carbon dioxide, and the intersegmental membrane between the fourth and fifth abdominal segments was cut by a razor blade. The battery was pushed into the opening using a tweezer so that the stimulation channels were positioned close to the opening. After that, the board was implanted between the second and third abdominal segments. Care was taken to keep the insect anesthetized and to gently insert the board and battery not to damage the internal tissues.

After surgery, individual cockroaches were kept in plastic containers and allowed to access food and water freely. The plastic containers were cleaned once a week to reduce the likelihood of infection and illness.

### 2.5. Video recording and analysis

Animal's behavior in the track was recorded from side view using a webcam (C920, Logitech) at 30 frames/s. The resolution of the video frame was 1920 x 1080 pixels. The track was illuminated by a custom LED array 100 cm above.

In each trial, the insect was released at the entrance of the track with the head directing toward the shutter and allowed to walk freely. If the insect failed to reach the shutter, i.e., contacting the shutter with any part of its body including antenna, within 10 s, it was collected and returned to the container. The insect was allowed to rest for at least 1 min before the next trial. Each trial ended when the insect 1) passed through the gap, 2) climbed over the shutter or the wall, 3) turned away from the shutter and tried to return to the entrance, or 4) 1 minute passed after the insect was released. Only such trials were accepted for analysis.

### 2.6. Electrode implantation and electrical stimulation

A Teflon coated silver wire (786000, AD Instruments) was soldered to each stimulation channel on the board, and a platinum wire was soldered on the other tip of the silver wire. The platinum wire was cut in 10 mm and only this part was used as an electrode. Two out of four electrodes were implanted into the left and right antenna, and one electrode into the cerci. The electrodes were fixed on antennae or cerci using melted beeswax and further secured with superglue. The cockroaches have cerci on both sides of the abdomen, the silver wire had an extra branch with a platinum tip. The last electrode was folded under the board and implanted with it into the abdomen.

The DAC provided a biphasic square pulse train to stimulate cockroaches. Amplitude was 2.5 V, pulse width was 12 ms, and the duration was 400 ms or 1200 ms.

### 2.7. Locomotion control and tracking

We conducted a point-to-point navigation to test the controllability of cyborg cockroaches. A custom-made GUI was developed to deliver stimulation to the insects based on an automatic navigation algorithm. The cyborg was released at the starting point on a flat arena and stimulated to reach the goal.

A 3D motion tracking system was used to measure the locomotion of the cockroaches during navigation. Six infrared cameras (T40 camera, VICON) were mounted on an aluminum frame to capture a 3D marker placed on a cyborg cockroach. The software (Vicon Tracker, VICON) obtained the position and orientation of the marker. The custom-made GUI linked with the tracking software and stored the locomotion of the cyborg and stimulation parameters.

### 2.8. Statistics

A Chi-square test and a one-way ANOVA was used to determine whether there are any significant differences between experiment conditions. For ANOVA test, individual cockroaches were considered as a random effect. All statistical analyses were performed on MATLAB software.

## 3. Result and Discussion

### 3.1. Miniaturize backpack design and fabrication

First, control backpack form factor design was carried out to enable full implantation of the backpack and the battery into the cockroach's abdomen. Wireless communication and multi-channel electrical signal generation are finalized as two main block functions of the backpack due to their critical impact on the study. In addition, the battery powering the backpack has its size and capacity constrained due to them being implantable inside the insect body. Hence, microcontroller CC1310F128 was chosen as the main processing unit for the backpack due to its small footprint of 4 mm x 4 mm and capability to perform wireless communication with ultra-low power consumption (Sub 1-GHz, transmitting at 10-dBm output power/ receiving current at 13.4 mA and 5.5 mA respectively) (Figure 1). Electrical signal for stimulation was generated by Digital-to-Analog Converter (DAC) Integrated Circuit (IC) AD5624 with 5 V reference voltage level. Comparing to other studies which either employed pulse-width modulation (PWM) signal directly from general purpose input/ output (GPIO) pins of the microcontroller [2] or featured DAC AD5504 (5 mm x 6.4 mm) [5,7], not only the stimulation generator retains the robustness in signal isolation, highly accurate and customizable stimulation signal generation but also comes with a significantly small footprint of 3 mm x 3 mm. Introducing castellated holes as debugging connector instead of standard configuration or pad connectors allows the control backpack to be more compact, while still secures its programmable capability. With the optimisation of components selection and layout design, the control backpack was successfully fabricated with dimension of 10 mm and 10 mm, which is suitable for full implantation inside the insect's body.

### 3.2. Obstacle negotiation

The cockroach released in the track approached the obstacle (shutter) and exhibited a complex sequence of behavior to negotiate it. To quantify the entire process of the behavioral sequence, we split it into smaller elements as below.

(A) Contact: Any part of the cockroach body contacted the shutter. The cockroach approached the obstacle (shutter) while sampling the space by antennae. Consequently, all trials started with antennal contact (Fig.2B).

(B) Tunnel: In this phase, the cockroach maintained a lower posture and pushed its head under the shield to go through the gap (Fig. 2B). It could push the shutter repeatedly or move left or right with rubbing the head on the shutter, however, such movements were considered to be included in single attempt of tunneling as far as the head remained under the shutter. The onset of tunnel was defined as the timing when the cockroach's head came out from the gap.

(C) Climb: The cockroach raised its body to climb up the obstacle. It put the forelegs (and sometimes middle legs) on the shutter or the wall, but the hind legs remained on the floor (Fig. 2B).

(D) Explore: This was defined as a transition phase and included two different cases: 1) the cockroach walked backward and pulled its head out from the shutter after an attempt of tunneling, and 2) the cockroach stopped leaning on the wall or shutter and descended to the floor.

(E) Pass: The cockroach went through the shutter and was free from it. We considered the cockroach completely passed the shutter when the 5$^{th}$ abdominal segment cleared it because the height of this part was smaller than 8 mm.

(F) Stuck: The cockroach stopped moving under the shutter. Any part of the body between the head and the 4$^{th}$ abdominal segment remained under the shutter.

(G) Return: The cockroach turned more than 90 degrees away from the shutter. No cockroach returned immediately after contacting the shutter.

(H) Exit: The cockroach completely left the floor and climbed over the shutter or the wall.

As the cockroach was given sufficient time to negotiate the obstacle, all trials ended in any of (E) Pass, (F) Stuck, (G) Return, or (H) Exit.

When the intact cockroach encountered the shutter, it predominantly chose to tunnel it (75 among 82 trials, 91%). The illumination and the obstacle height, as shown in the past studies [8]. Compared to the approach phase, the insect lowered its body while tunneling. With this posture, the cockroach could move its antennae behind the shutter and detect the presence of an empty space (Fig. 2B). As the gap under the shutter was smaller (8 mm) than the insect's body height, the cockroach was first blocked by the shutter to clear it (Supplementary movie 1 and Supplementary Fig. 2). Most of the intact cockroaches successfully passed through the shutter by lifting it to enlarge the gap (74 among 77 attempts, 96%). When the backpack set was mounted on the cockroach, it struggled to go into the gap. (Fig. 3A, B). Similar to intact cockroaches, most of the cockroaches tried to tunnel under the shutter after contacting it. Although the cockroach turned down its head to go under the shutter, the backpack disturbed smooth transition to tunneling. To break into the gap, the cockroach needed to lift the shutter much higher than its body height repeatedly (Fig. 3B and Supplementary movie 2). The cockroaches often gave up tunneling it and proceeded to the 'explore' phase followed by returning or climbing (Supplementary movie 3). As a result, only 24 out of 71 attempts at tunneling (18%) were successful (Fig. 4A). Most tunnel attempts (48%) ended in 'stuck', indicating the cockroach settled down at the shutter or failed to pass through it within 1 minute. Meanwhile, the backpack set did not have noticeable influence on climbing (Fig. 3C). Implantation of the backpack greatly improved the gap negotiation (Fig. 4B and Supplementary movie 4). The implanted cockroaches also showed a strong tendency to tunnel under the shutter (37 out of 41 attempts), but 90% were successful. Half of the rest got stuck at the shutter and the other proceeded to 'explore' phase. No implanted cockroach chose to return as the intact cockroaches (Fig. 4B).

To examine if the successful rate of tunneling differed between three groups, we conducted a statistical test (Fig. 4C). If a tunneling attempt ended in pass, we counted it as success. On the other hand, 'explore' and 'stuck' were considered failures. There was significant difference among group (Pearson's chi-square test, *p < 0.01*). A post-hoc test detected significant difference between intact and mounted group (*p < 0.01*), and between mounted and implanted group (*p < 0.01*), whereas there was no difference between intact and implanted group (*p = 0.20*).

We measured the time that the cockroach spent passing through the shutter as traversal time (Fig. 4D). The traversal time was defined as the time from the onset of 'tunnel' and 'pass'. The intact cockroach spent 11.77 ± 1.37 s (mean ± SE, N = 3, n = 74). The mounted cockroach required longer time (20.60 ± 3.16 s, N = 3, n = 13), whereas the implanted cockroach passed with shorter time (8.81 ± 1.00 s, N = 3, n = 37). The traversal time was significantly difference among group (one-way ANOVA, *F = 6.3876, p < 0.01*). A post-hoc test detected significant difference between intact and mounted group (*p < 0.01*), and between mounted and implanted group (*p < 0.01*), whereas there was no difference between intact and implanted group (*p = 0.61*).

These results suggested that 1) arrangement of the backpack did not alter the innate tendency of the cockroach to pass through a narrow gap, and 2) compared to the conventional arrangement, i.e., attaching the backpack on the thorax, implantation significantly improved the traversability of the cockroach.

### 3.3. Locomotion control of cyborg cockroach

In the above section, we showed that our implantation method has a potential advantage for outperforming the conventional preparation method, i.e., mounting on top. To address this, we examined the locomotion control of cyborg insects with implanted backpack (Fig. 5). It is well known that the cockroach initiates walking or accelerates when electrical stimulation is delivered to cerci and shows left/right turning to right/left antenna stimulation. Using this protocol, the cyborg cockroach was stimulated to navigate from the virtual starting point to the virtual target (Fig. 5A). The automatic navigation algorithm stimulated cerci for acceleration when the cyborg cockroach was not walking. On the other hand, the algorithm stimulated left antenna when the heading of the cyborg was biased larger than 45 degrees to the left from the target and *vice versa*. During the navigation, cyborg cockroach could spontaneously turn to left or right, and then, it received antenna stimulation to turn back to the target. As a result, 33 out of 35 trials, our cyborgs reached within 50 mm from the target (94.3% N = 3). The reaction to each type of stimulation confirmed the faithful locomotion control of the cyborg cockroach (Fig. 5D-F). The mean turning velocity for left stimulation (right antenna was stimulation) was 24.26 ± 20.41 degrees/s (mean ± SE, N = 3, n = 144) whereas -23.45+ 17.51 degrees/s for right stimulation (mean ± SE, N = 3, n = 91). Because the navigation algorithm provided electrical stimulation to cerci when the cyborg cockroach stopped walking, the mean walking velocity before stimulation was close to zero (Fig. 5E). The cockroach was induced to initiate walking after stimulation. The forward velocity decreased after stimulation but remained over zero, indicating the cyborg cockroach continued to walk forward. The mean forward velocity during 1 s after stimulation was 33.01 ± 13.77 mm/s (mean ± SE, N = 3, n= 39).

As shown in Section 3.2., the cockroach could walk backward when it was blocked by an obstacle. Inspired by this, we examined the locomotor reaction to electrical stimulation to both antennae (Fig. 6). The short stimulation (400 ms) passed through both antennae induced a reduction in forward velocity (Fig. 6A). However, the mean forward velocity during 1 s after stimulation was 3.16 ± 2.20 mm/s, indicating the cyborg moved forward in total. Longer stimulation (1200 ms) elicited prolonged reaction, resulting in the mean forward velocity –2.03 ± 2.50 mm/s (Fig. 6B).

### 3.4. Navigation in a concealed environment

We presented that our implantation method improved the traversability of cyborg insect in an obstructed terrain and that locomotion of the implanted cyborg cockroach was well-controllable. Therefore, we conducted a conceptual demonstration where a cyborg cockroach trapped in a concealed space navigated another space through a narrow gap in the environment (Fig. 7 and Supplementary movie 5). At the beginning, the cyborg cockroach stood at the starting point in a small space that had no major entrance or exit. The cyborg was stimulated to approach the air vent like structure on the wall. As the cockroach had an innate tendency to go into a narrow space, it stepped on the structure and pushed its head in the slit on it without stimulation. The cockroach came out from the space, crossed the main space at the center, and came near to a small space on the other corner. The same structure on the wall allowed the cockroach to go through the gap to reach the target.

### 3.5. Impact of implantation

We surveyed the survivability of the cockroaches after the backpack implantation. Four out of seven cockroaches survived for 1-week post-implantation, whereas three died within 6 days. Because the Madagascar hissing cockroaches live longer than 2 years, this survival period was quite short, leading to the estimation that the implantation gave a lethal damage on the cockroaches. Insects including cockroaches have a tubular heart on the dorsal side, and the heart is the potential organ which may contribute to this short survival period (Supplementary Table 1). In 3 cockroaches heart activities became invisible after surgery, and all these cockroaches died shortly. On the other hand, 3 of 4 implanted cockroach showed obvious heart activities after implantation, although the frequency was lower than the one before implantation., This implied that the implanted backpack caused disability in the heart and triggered the death of the cockroaches.

### 3.6. Future improvement

Despite successfully demonstrating insect navigation control in cramped environment, there is room for improvement for machine-insect fusion. First, the electronic system has potential to come with even more compact sizes. This is feasible by combining all main IC (microcontroller, DAC, power management unit, passive component if applicable) into a single System-on-Package (SoP). In addition, replacing rigid printed circuit board material and components by flexible alternatives such as soft bioelectronic devices [9] will allow the control system to fit with the curved form of insect abdomen cuticle; hence, the compatibility of machine and insect parts is enhanced. Furthermore, improvement on power system can be accomplished by replacing LiPo battery by bio-fuel cell [10, 11] or solar cell [6], subjected to operation condition. Such methods reduce the need for invasive operation and lead toward a harmonic fusion of machines and insects.

### 4. Conclusions

By comparing two different arrangements of the board and battery, we demonstrated the streamlined shape facilitated gap negotiation of the cyborg cockroach. The board used in this paper, 4 mm high corresponding about one-third of the intact body height, reduced the success rate of gap negotiation to 18% when it was mounted on the body. Meanwhile, 90% of cockroaches with implanted board successfully went through the gap. As the cyborg with implanted board remained well-controllable, we could navigate them from point to point through a narrow gap like an air vent cover. Our cyborgs promote application in human living environment.

**Author contribution**

K.K. and H.S. conceptualized the research. L.D.L. designed the board and developed the firmware. K.K. developed the navigation algorithm. K.K. conducted all experiments and data analysis. K.K., L.D.L., and H.S. wrote the paper.

**Acknowledgment**

The authors thank P. Thanh Tran-Ngoc at School of MAE, NTU for their helpful comments and advice. The authors offer their appreciation to Ms. Kerh Geok Hong, Mr. Tan Kiat Seng, and Mr. Roger Tan Kay Chia at School of MAE, NTU for their continuous support in setting up and maintaining the research facilities. The authors received funding from JST (Moonshot R&D Program, Grant Number JPMJMS223A).

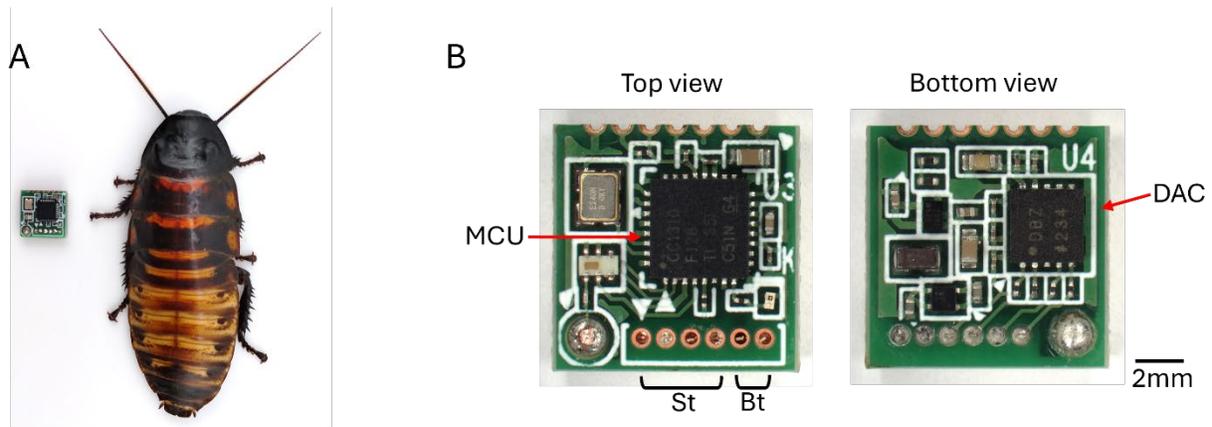

**Fig. 1. Design of backpack.** A. Top view of Madagascar hissing cockroach and the developed backpack. B. Top and bottom side of the backpack. MCU: microcontroller unit, DAC: digital-to-analog converter, St: stimulation channel, Bt: battery terminal

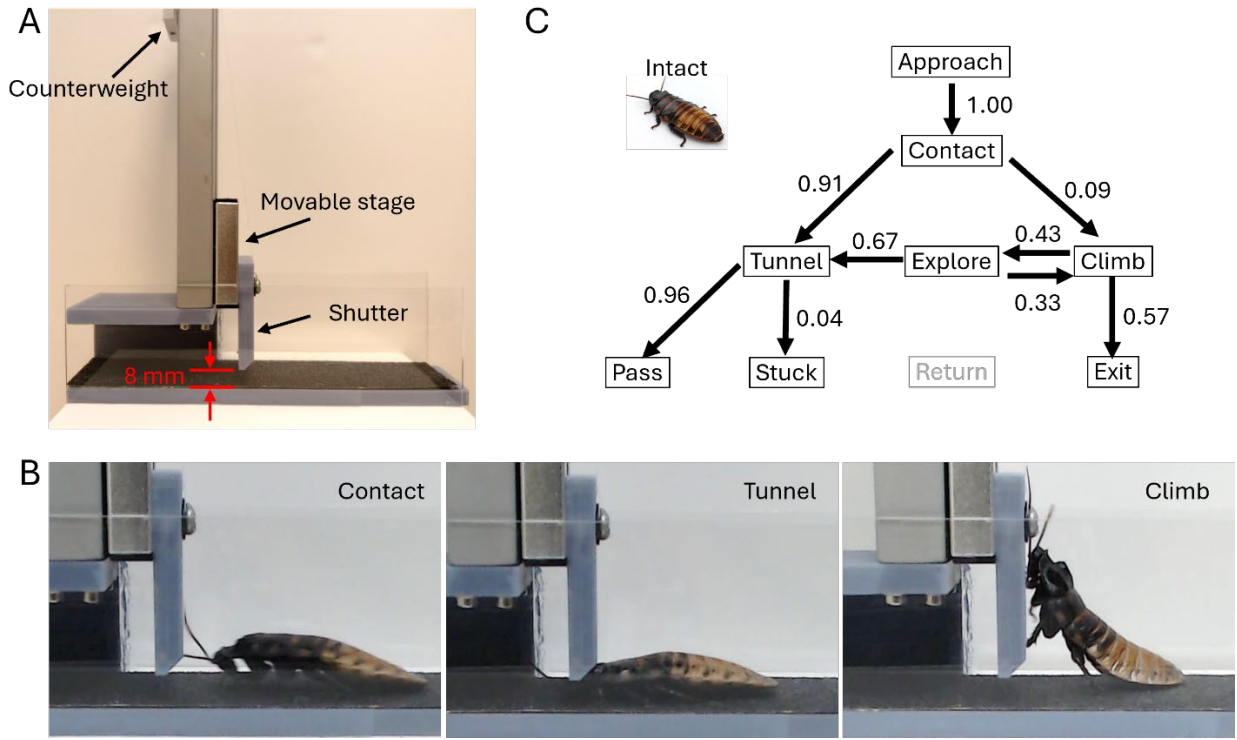

**Fig. 2. Obstacle negotiation of intact cockroach. A.** Apparatus. **B.** Snapshot of cockroach during antennal contact, tunneling, and climbing. **C.** Block diagram of cockroach's behavior.

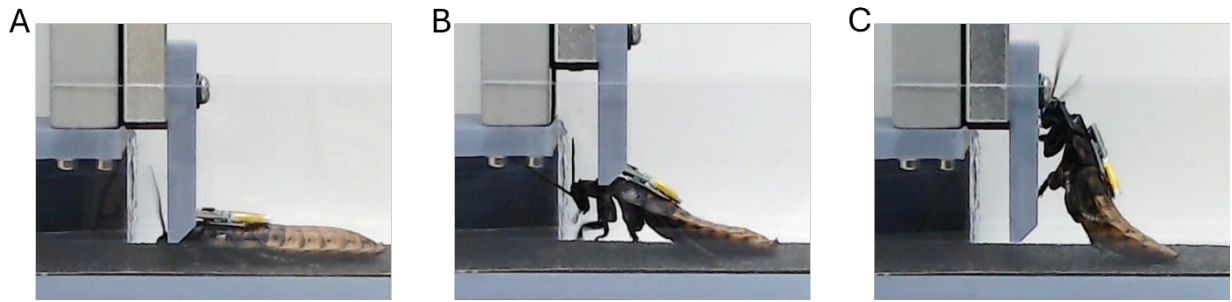

**Fig. 3. Backpack impacts on gap negotiation. A.** snapshot of cockroach during tunnel phase. **B.** Cockroach lifting the shutter. C. snapshot of cockroach during climbing.

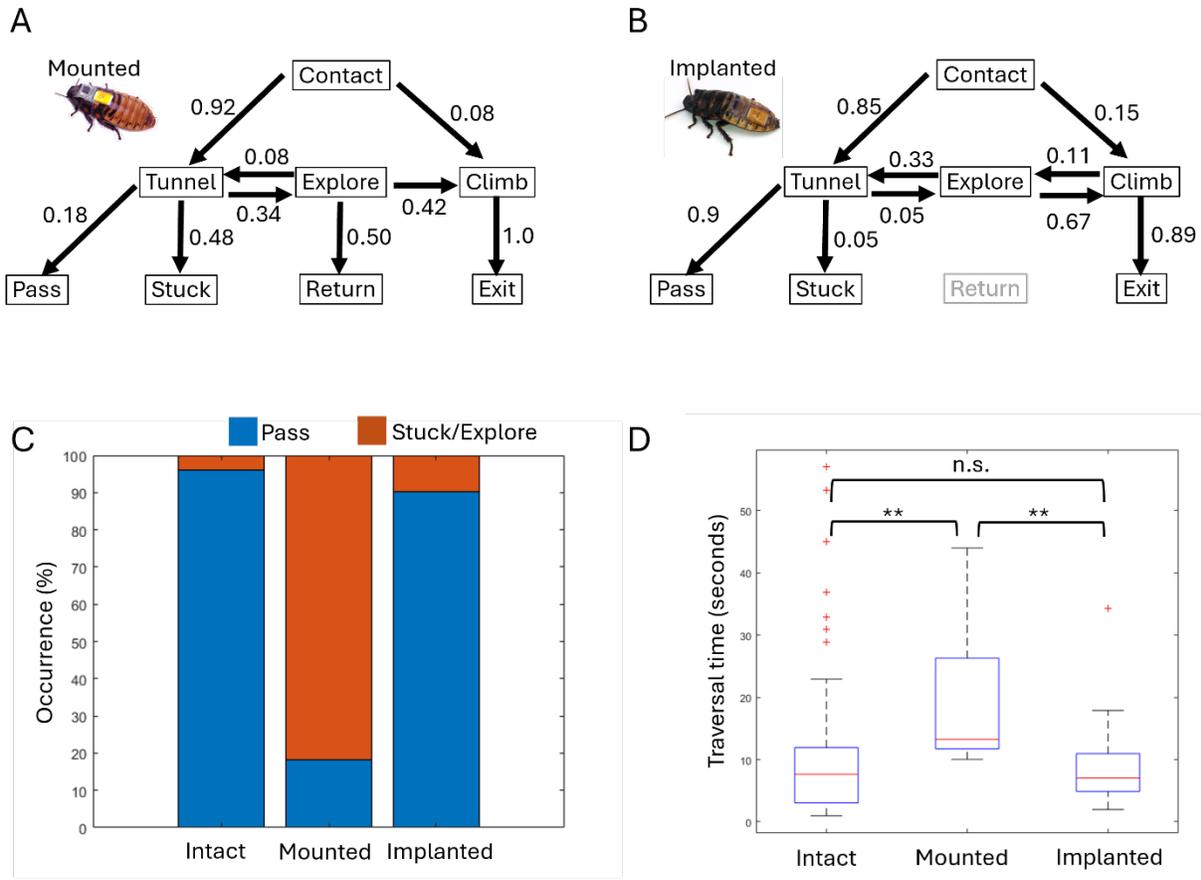

**Fig. 4. Obstacle negotiation of mounted and implanted cockroach. A.** Block diagram of the behavior of a mounted cockroach. **B.** Block diagram of the behavior of implanted cockroach. **C.** Success rate of 'tunnel' behavior. **D.** Traversal time.

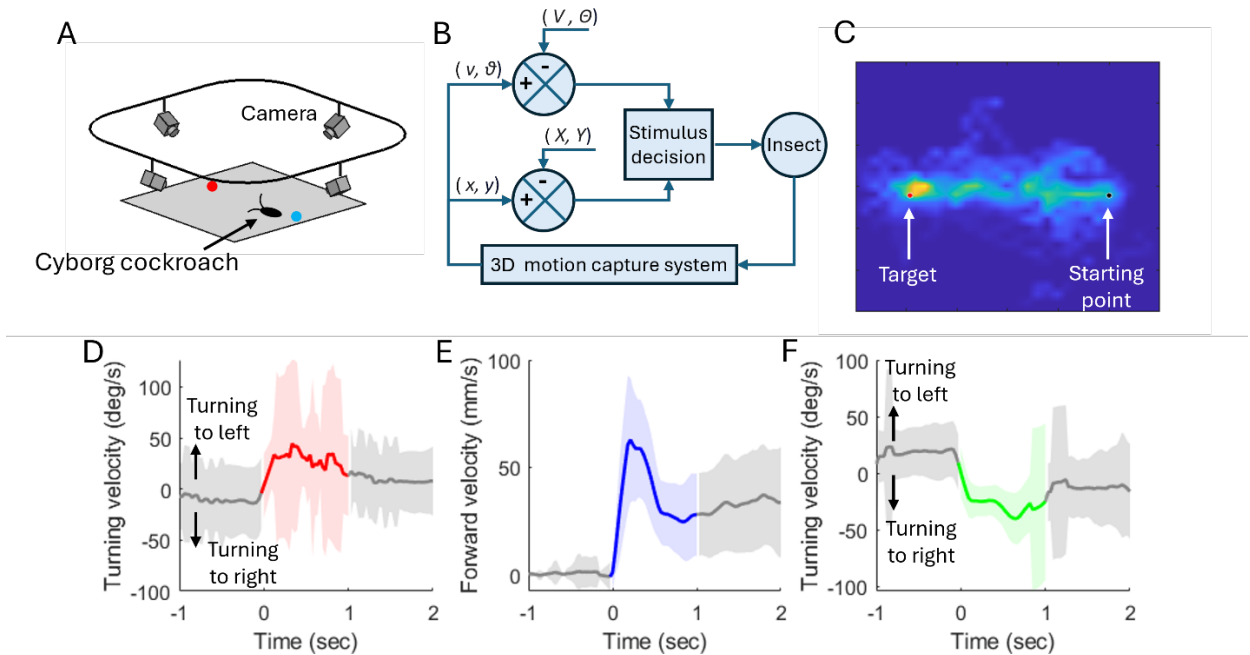

**Fig. 5. Locomotion control of cyborg insect. A.** Schematic view of experiment setup. Blue circle: virtual starting point. Red circle: virtual target. **B.** Block diagram of automatic navigation algorithm. X: x-coordinate of the target. Y: y-coordinate of the target. x: x-coordinate of the cyborg. y: y-coordinate of the cyborg. V: threshold for forward velocity. Θ: threshold for antenna stimulation. v: forward velocity of the cyborg. θ., turning velocity of the cyborg. **C.** Density map during navigation. **D-F.** Turning and forward velocity of the cyborg. The thick lines indicate the mean value and shades indicate S.D.

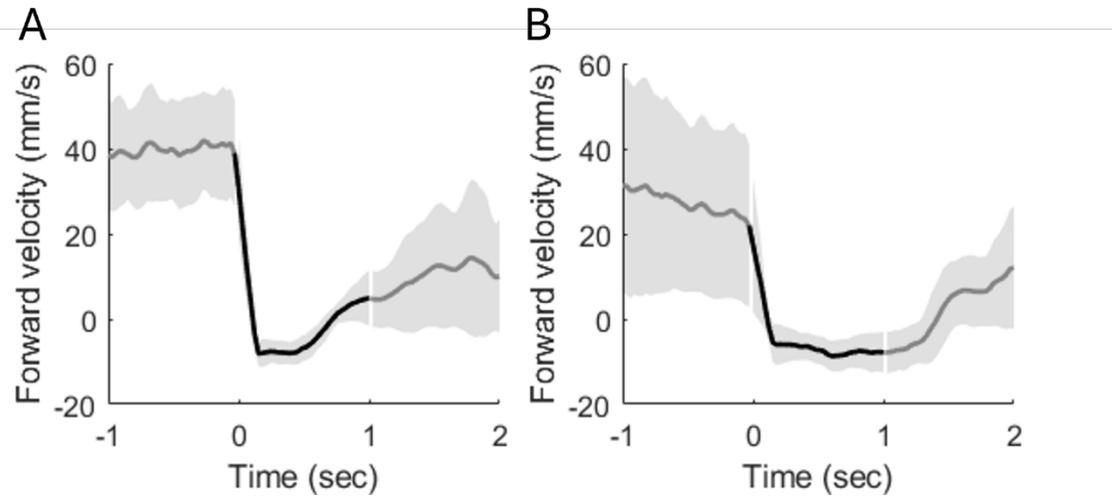

**Fig. 6. Electrical stimulation to both antennae induced backward waking. A.** Forward velocity during 400 ms stimulation. **B.** Forward velocity during 1200 ms stimulation.

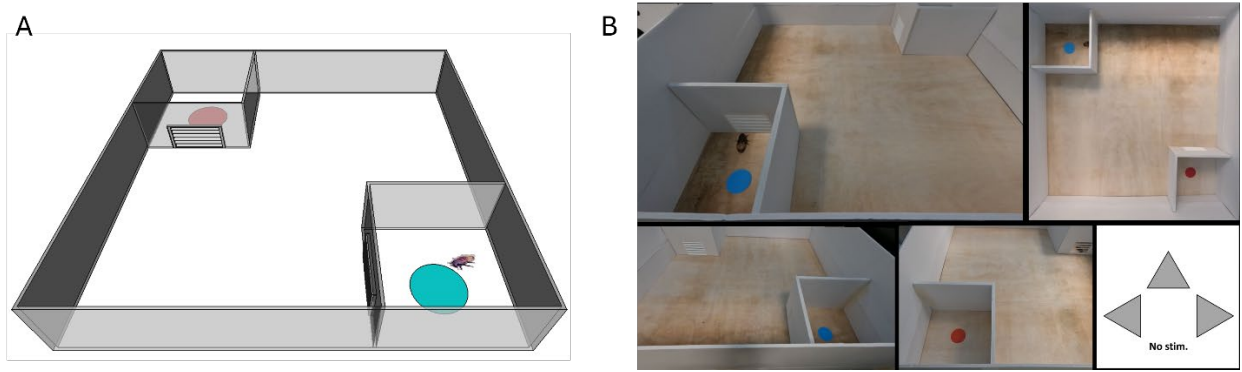

**Fig. 7. Design of conceptual demonstration.** A. Schematic view of the arena. Green circle: starting point. Red circle: target. B. Snapshot of the video. The locomotion of the cyborg cockroach was filmed from above, left, right, and front.

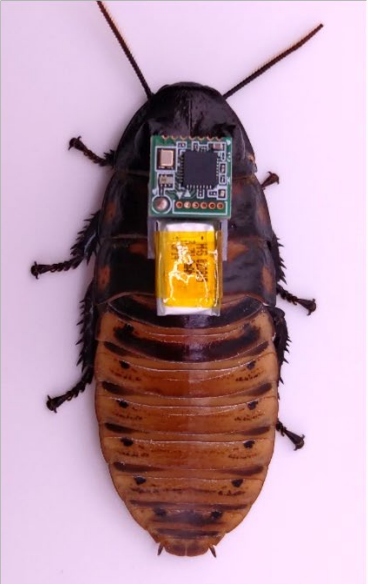
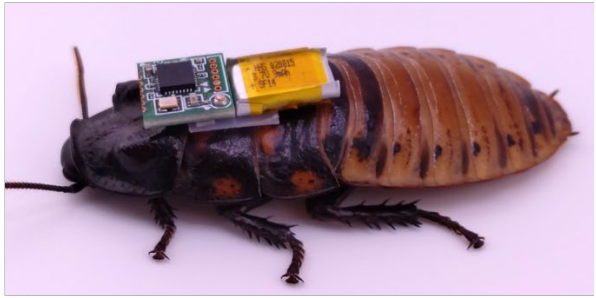
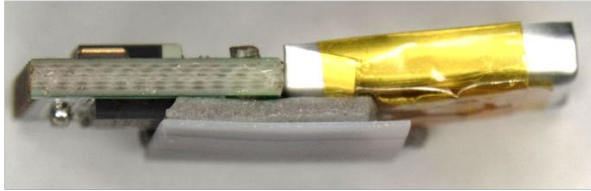

**Supplementary Fig. 1. A.** Top view of the cockroach. The backpack and battery was attached to the mesothorax. **B.** Side view of the cockroach in A. **C.** Side view of the backpack set.

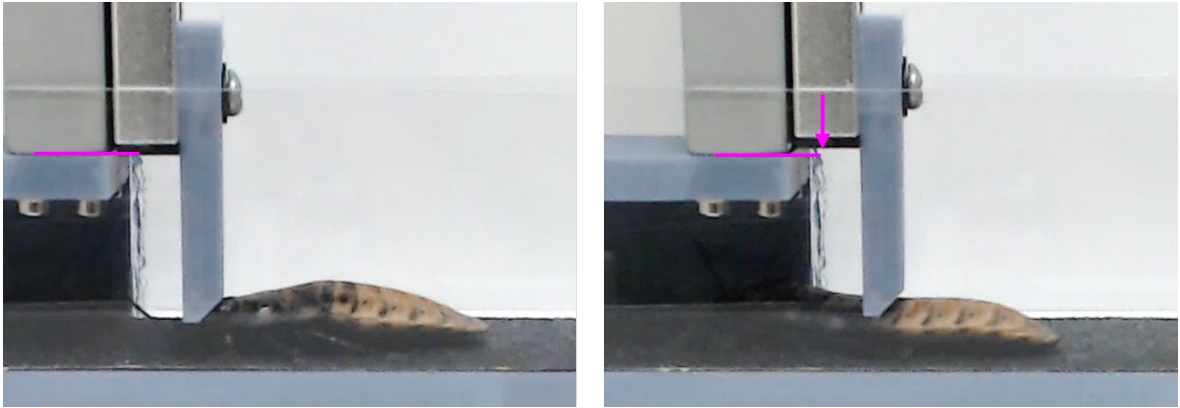

**Supplementary Fig. 2.** Cockroach lifting the shutter. Left: cockroach in 'tunnel' phase. Right: cockroach passing through the gap by lifting it. The level of horizontal magenta line in each picture indicates the same height. Note that there is a small gap between the magenta line and the bottom edge of the stage (right, arrow).

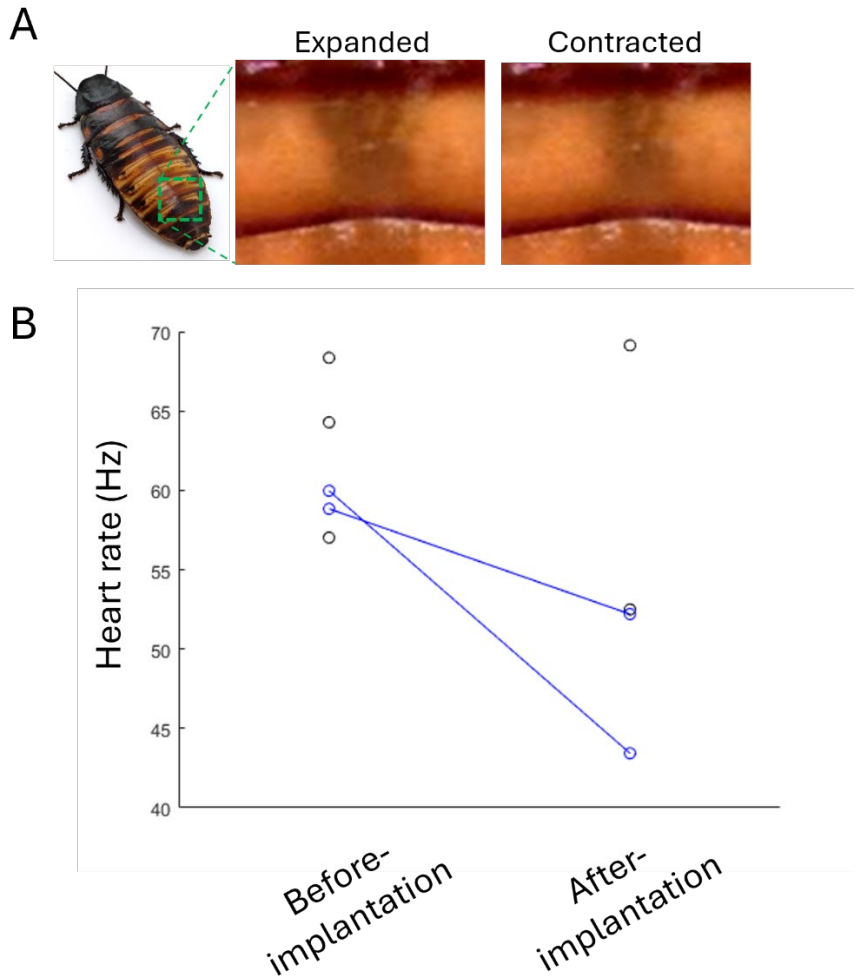

**Supplementary Fig. 3.** Comparison of heart rate before and after backpack implantation. **A.** Heart rate measurement. Heart activities were measured from a cockroach in quiescence. The live insect was placed in a plastic dish (90 mm in diameter and 10 mm in height) with dorsal side up. A spacer was positioned on the head, and the cockroach was gently pressed with a transparent lid. The lid was fixed using 2 pieces of masking tape to immobilize the insect and the spacer. The spacer shaded illumination from the surroundings to make the insect quiescent in the dish and also induced body stretching on the cockroach. The heart activities were observed at the 6$^{th}$ abdominal segment and recorded as a video using a digital microscope (VHX-7000, KEYENCE). The video was recorded for 1 minute with a resolution n of 800 x 640 pixels and a frame rate of 30 frames/s. The heart rate was calculated as: Heart

Rate = (duration between the first and last heart beating)/(n-1), where n was the number of beating in 1 minute. **B.** Heart rate before and after implantation.

| Animal | Survival period (days) | Heart rate (Hz) | |
|---|---|---|---|
| | | Pre-implantation | Post-implantation |
| 1 | 5 | 64.27854034 | not measurable |
| 2 | 9* | 58.83364872 | 52.20010038 |
| 3 | 4 | 68.34656541 | not measurable |
| 4 | 6 | 57.01679959 | not measurable |
| 5 | 7 | 59.96950703 | 43.43329886 |
| 6 | 17* | N.A. | 69.12910878 |
| 7 | 53* | N.A. | 52.49343832 |

**Supplementary Table 1.** Survival after implantation and heart rate. NA: heart rate was not measured. Asterisk indicates the insect is alive when writing.

**Supplementary Movies** are available here;
https://www.dropbox.com/scl/fo/46t8cv7vae966so5o0xu1/ANkZxjGXDi9e9B0f9DIHlU8?rlkey=ivlh688e2v41q2zmsgr6rc0hy&dl=0